\documentclass[sigconf]{acmart}

\def\BibTeX{{\rm B\kern-.05em{\sc i\kern-.025em b}\kern-.08emT\kern-.1667em\lower.7ex\hbox{E}\kern-.125emX}}
    
\copyrightyear{2019}
\acmYear{2019}
\setcopyright{acmlicensed}
\acmConference[ICONS '19]{ICONS '19:International Conference on Neuromorphic Systems}{July 23--25, 2018}{Knoxville, TN}
\usepackage[compact]{titlesec}
\titlespacing{\section}{0pt}{1ex}{1ex}

\setcopyright{none}
\renewcommand\footnotetextcopyrightpermission[1]{}
\begin{document}

%
\title{Neuromorphic Architecture Optimization for Task-Specific Dynamic Learning}

%
\author{Sandeep Madireddy}
\email{smadireddy@mcs.anl.gov}
\affiliation{%
  \institution{Mathematics and Computer Science Division, Argonne National Laboratory}
  \streetaddress{9700 S. Cass Avenue}
  \city{Lemont}
  \state{Illinois}
  \postcode{43017-6221}
}

\author{Angel Yanguas-Gil}
\email{ayg@anl.gov}
\affiliation{%
  \institution{Applied Materials Division, Argonne National Laboratory}
  \streetaddress{9700 S. Cass Avenue}
  \city{Lemont}
  \state{Illinois}
  \postcode{43017-6221}
}

\author{Prasanna Balaprakash}
\email{pbalapra@mcs.anl.gov}
\affiliation{%
  \institution{Mathematics and Computer Science Division, Argonne National Laboratory}
  \streetaddress{9700 S. Cass Avenue}
  \city{Lemont}
  \state{Illinois}
  \postcode{43017-6221}
}



 




%
\renewcommand{\shortauthors}{Madireddy, Yanguas-Gil and Balaprakash}

%
\begin{abstract}

The ability to learn and adapt in real time  is a central feature of
biological systems. Neuromorphic architectures demonstrating such versatility
can  greatly enhance  our ability to efficiently process information at the
edge. A key challenge, however, is to understand which learning rules are best
suited for specific tasks and how the relevant hyperparameters can be
fine-tuned. In this work, we introduce a conceptual framework in which the
learning process is integrated into the network itself. This allows us to cast
meta-learning as a mathematical optimization problem. We employ DeepHyper, a
scalable, asynchronous model-based search, to simultaneously optimize the
choice of meta-learning rules and their hyperparameters. We demonstrate our
approach with two different datasets, MNIST and FashionMNIST, using a network
architecture inspired by the learning center of the insect brain. Our results
show that optimal learning rules can be dataset-dependent even within similar
tasks. This dependency demonstrates the importance of introducing versatility
and flexibility in the learning algorithms. It also illuminates experimental
findings in insect neuroscience that have shown a heterogeneity of learning
rules within the insect mushroom body.

\end{abstract}

%
%


%

%

%
\maketitle
\vspace{-0.15in}
\section{Introduction}

The challenges of designing smart systems for 
edge-processing applications are different from those of
conventional machine learning (ML) approaches. Usually,
ML approaches rely on
large datasets and highly optimized training algorithms. The
resulting architectures are specific to certain tasks, and their parameters are traditionally
fixed once the training is over. In contrast, in edge-based applications, devices in the field should be capable of learning and responding quickly, often from noisy and incomplete data, and should be capable of adapting to unforeseen changes in a safe, smart way. 

A promising approach for dynamic learning is to develop systems that are based on biological systems. 
In particular, biological neural networks are both
dynamic and plastic, with the capability to
incorporate multiple functionalities depending
on the context, and can carry out context and task-dependent learning. Insects in particular are ideal model systems for edge-processing applications: they display impressive capabilities despite their central neural system being composed of a small number of neurons, 100,000 in the case of Drosophila melanogaster and 1,000,000 in
the case of the honeybee. 
Recently, researchers have expressed renewed interest in dynamic learning, but the focus has been on spiking neurons. Several works have explored ways in which Hebbian-inspired rules, primarily based on spike-timing-dependent plasticity, can be used to implement learning in spiking networks~\cite{fremaux_neuromodulated_2016}.
In a broader sense, however, dynamic learning has deep roots in artificial neural networks; and the question of batch size and its impact on a trained network's ability  to learn and generalize is still an open research question.

Here we focus on this problem from a different perspective: learning can be viewed as a dynamic process that alters the network itself as it processes and assigns valence to certain inputs and create associations over time. The question then is, How can we find the optimum architecture and learning rules for a given task? Evolution has pushed biological systems to find highly efficient solutions within biochemical constraints. We seek to develop an analogous approach to help us identify optimal architectures for dynamic neural networks.

To accomplish this goal, we have implemented networks capable of dynamic learning as recurrent neural networks where plastic synaptic weights are treated as layers of the network. We parameterize the learning rules and expose them as hyperparamters. Doing so allows us to approach the problem of designing an effective dynamic learning from an optimization perspective. We then employ a scalable, asynchronous model-based search (AMBS) algorithm to simultaneously optimize learning rules and their hyperparameters. We adopt the AMBS implementation in  DeepHyper~\cite{Balaprakash_DH_2018}, a scalable optimization package that is built to take advantage of leadership-class computing systems through parallel algorithms and efficient workflow management systems and hence provides advantages in speed to solution in addition to accuracy.
To this end, we make the following contributions:
(1) a new approach to designing and optimizing neuromorphic architectures for task-specific dynamic learning by combining asynchronous model-based search with modulated learning and
(2) demonstration that task-specific meta-learning rules are essential to extract the maximum performance from the learning model using two widely used ML benchmark problems in combination with mushroom body architecture.
\vspace{-0.03in}
\section{Dynamic Learning}
\vspace{-0.05in}
The architectures and meta-learning rules considered in this study are described below
\vspace{-0.10in}
\subsection{Architecture}
\begin{figure}[th!]

    \includegraphics[width=0.8\linewidth]{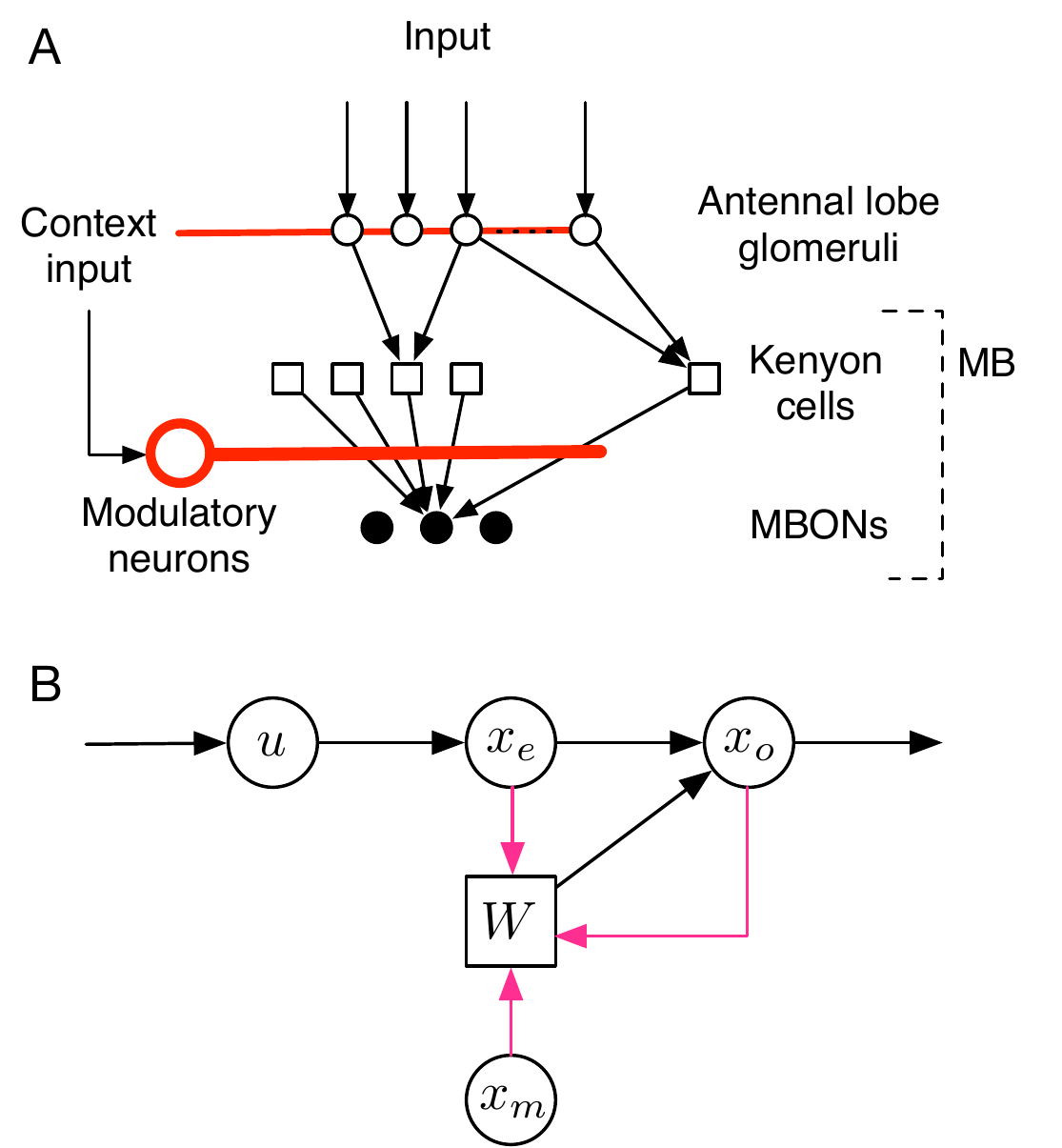}\\
    \caption{(A) Architecture inspired by
    the insect mushroom body: input information
    is sparsely fanned out from the
    antennal lobe to the Kenyon cells and then
    densely fanned into the output neurons.
    Modulatory neurons provide contextual
    information to carry out both context-dependent filtering and context-dependent learning. (B)
    Simple scheme of a recurrent implementation of dynamic learning via local learning rules: synaptic weights are treated as a separate layer of the network, receiving inputs from the presynaptic, postsynaptic, and modulatory layers.}
    \label{fig:scheme}
\end{figure}

\vspace{-0.05in}
To explore architecture optimization for dynamic
learning applications, we consider architectures that are inspired by the insect brain and, in particular, one
of its key centers for olfactory (and, in hymenopterans, visual) memory: the mushroom body. The mushroom body
constitute the third and fourth layers of olfactive processing in insects: sensing information is pooled in
a first layer called the antennal lobe and then sparsely projected into the Kenyon cells in the mushroom body; see Figure \ref{fig:scheme}(a). This
projection creates a sparse representation of the input that helps enhance its dimensionality. This layer is then
densely connected into a few output neurons, which are recurrently connected. 
A key aspect of the mushroom body is that learning takes place primarily in this last step. Moreover,
modulatory neurons innervating the mushroom body control when and where learning takes place\cite{aso_neuronal_2014, hige_heterosynaptic_2015}.

In this work, we have abstracted this architecture to explore the optimization and accuracy of different dynamic
learning algorithms based on modulated learning and local learning rules. Since learning takes place primarily in 
the last layer, we can model the input layers as a functional transformation of our input, so that $\mathbf{x}_e = f(\mathbf{u};
\mathbf{,})$. Here $\mathbf{u}$ is our input, $\mathbf{x}_e$ represents the population of Kenyon cells in the mushroom
body, and $\mathbf{m}$ is a vector of modulatory interactions providing a
context-dependent input for context-dependent filtering of the input stream. Such context-dependent processing
has been experimentally observed in the first layer of olfactive processing of insects. The output layer takes a linear combination of inputs that are densely connected with recurrent, cross-inhibitory 
interactions.

In addition to this processing component, we have a set of modulatory neurons  $\mathbf{m}$ providing context-dependent
information. This can be externally determined either for supervised learning or as a reinforcement signal, and their activity can
be modified by the output of the neurons or by an independent
modulatory component in our network, either as feedback or for cases where the system has internally to decide among multiple tasks.

Our goal is to explore dynamic learning using local learning rules. Instead of considering the network and the
 training algorithm as two separate entities, we consider that the evolution of the synaptic weights is given by the following general rule.
\begin{equation}
\label{eq_rule}
\dot{\mathrm{W}} = f\left(\mathbf{x}_e, \mathbf{x}_o, \mathbf{x}_m, \mathrm{W};\mathbf{\beta}\right)
\end{equation}
Here $\mathbf{x}_e$,  $\mathbf{x}_o$, and $\mathbf{x}_m$ generally represent the activity of
presynaptic, postsynaptic, and modulatory neurons, respectively, and $\beta$ represents a set of hyperparameters controlling how learning
takes place.

With this approach, the system's ability to dynamically
learn depends both on the selection of the actual learning
rule and on the
hyperparameters involved. To better understand how
different learning rules affect the system's ability to dynamically
learn different tasks,
we propose a conceptual framework in which the learning
process is integrated into the network itself. If we 
treat the synaptic weights following Eq. \ref{eq_rule} as
first-class citizens of our network, we now have a recurrent
network that essentially modifies itself based on the
input and an external context, as shown in Figure \ref{fig:scheme}(b) .
The key advantage of this approach is that we can leverage
the capabilities of existing ML frameworks as well as 
existing approaches for network architecture optimization.

\subsection{Learning and Meta-Learning Rules}
\label{sec:meta_def}

Many dynamic learning rules have been discussed
in the literature in the context of unsupervised, supervised,
and reinforcement learning applications. Here we considered
a subset of these rules and modified some of them
to incorporate the presence of modulatory interactions regulating
when and where learning takes place.

We have explored the following meta-(local)learning rules:
 \begin{itemize}
 
 \item Modulated covariance rule (MCR):
 \begin{equation}
 \Delta W = \alpha \mathrm{ReLU}(x_m-x_o) x_e \left( x_m-\beta_1\right)
 \end{equation}

\item Nonlocal, stabilized covariance rule (NSCR):
\begin{equation}
 \Delta W = \alpha \sum_{m,o} \mathrm{ReLU}(x_m-x_o) x_e \left( x_m-\beta_1 W \right)
 \end{equation}

\item Nonlocal, stabilized correlation rule (NSCoR):
\begin{equation}
 \Delta W = \alpha g  \left( x_e x_m-\beta_1 W \right),
 \end{equation}
 where $g =  \sum_{m,o} \mathrm{ReLU}(x_m-x_o)$

\item Modulated Oja's rule (MOR):
\begin{equation}
 \Delta W = \alpha g  \left( x_e x_m-\beta_1 x_o^2 W \right),
 \end{equation}
 where $g = \mathrm{ReLU}(x_m-x_o)$.

\item Least mean square rule (LMSR):
\begin{equation}
 \Delta W = \alpha x_e( x_m-x_o)
 \end{equation}

\item Self-limited rule (SLR):
 \begin{equation}
 W =\frac{W + W_0\alpha g x_e}{1 + \alpha g (\beta_1 + x_e),}
 \end{equation}
where $g =  \mathrm{ReLU}(x_m-x_o)$

 \item General modulated rule (GMR):
 \begin{equation}
 \Delta W = \alpha x_m \left(\beta_1 x_o  + \beta_2(x_o-x_e) + \beta_3 \right)
 \end{equation}

\item General unsupervised rule (GUR):
 \begin{equation}
\Delta W = \alpha \left(\beta_1 x_o  + \beta_2(x_o-x_e) + \beta_3 \right)
 \end{equation}
\end{itemize}

In all these examples $\alpha$ is a parameter that controls
the overall learning rate.

Both MCR and NSCR are learning rules where the change in
the synaptic weight is proportional to the presynaptic
input $x_e$ and to the difference between the modulatory input
$x_m$ and the activity of the output neuron $x_o$. This provides
a natural feedback loop that interrupts learning once the desired
output is achieved. In both cases, the sign of the weight change is determined
by the difference between the modulatory input $x_m$ and a
threshold variable. The key difference between MCR and NSCR is 
that in the former learning takes place solely whenever the 
modulatory neuron is active, whereas in the latter a general
modulation term is applied to all output neurons. Also, in the
case of NSRC the loss term is proportional to the synaptic
weight to ensure the stability of the learning rule. 

NSCoR is similarly a normalized covariance term, where the 
synaptic weight increase is driven by the covariance between
the pre- and postsynaptic activity $x_e$ and the target
modulatory output $x_m$. 

MOR is simply a modulated version of Oja's learning rule, whereas
LMSE can be derived from a gradient descent rule
from a cost function equal to the mean square error
between the actual output and the expected output.

SLM is a self-limited rule in which synaptic weights are 
allowed to change only between 0 and $W_0$. The current expression
has been obtained from a fully implicit discretization of the
corresponding differential equation.

GMR and GMU are two general rules based on the activity
of pre- and postsynaptic neurons. In the former case the learning
rate is modulated by the difference between the expected and
the actual output of the network, whereas in the latter case
the evolution of the synaptic weight is
fully unsupervised.
We expect that GMR and GMU will perform better when the
modulatory neurons input the postsynaptic neurons. They will
also be sensitive to the cross-inhibition between the output layer,
which provides a natural competition between the output neurons. 

\section{Architecture Optimization}

We adopt a parallel asynchronous model-based search~\cite{Balaprakash_DH_2018} to learn
the optimal meta-learning rule and its corresponding hyperparameters for a given 
architecture and dataset combination. Parallelization of the parameter configurations 
evaluation is critical for scaling the optimization algorithms to handle architectures that
are computationally intensive to evaluate or have a large number of tunable parameters.

In the AMBS approach adopted, a surrogate model is fit between model parameters (such as the 
meta-learning rule choice and the hyperparameters within the rule) and validation accuracy 
(for a ML task on a given dataset) to guide the search. This surrogate is updated 
dynamically (and asynchronously) during each iteration of the search process by 
including the newly evaluated configurations, but without waiting for evaluations from 
all the active processes to be complete. This updated surrogate model is used to obtain 
promising configurations to evaluate for the next iteration. 

Crucial to this approach are the choice of the surrogate model and the criterion used to choose the promising configurations (acquisition function). We use a random forest regressor~\cite{breiman2001random} as the surrogate model since our search space consists of categorical parameters (choice of the meta-learning rule) and continuous variables (parameters inside the learning rule). Random forest is an ensemble machine learning approach that uses bootstrap aggregation (or bagging), wherein an ensemble of decision trees are  combined to produce a model with better predictive accuracy and lower variance. The acquisition functions originate from the Bayesian optimization literature~\cite{shahriari2016taking} and are used as strategies to balance exploration and exploitation in the search space. We use a hedging strategy~\cite{hoffman2011portfolio}, wherein at each iteration, the algorithm chooses from a portfolio of acquisition functions based on an online multiarmed bandit strategy.

\section{Results and Discussions}
We adopt two different datasets to study the hypothesis that task-specific meta-learning rules need to be considered and optimized in order to obtain the best predictive accuracy.  

The first dataset is the widely used benchmark  MNIST (Modified 
National Institute of Standards and Technology)~\cite{lecun1998gradient}, which consists of 
grey-scale digital images of handwritten digits (0--9) that have been hand labelled. This dataset comprises 60,000 training data and 10,000 testing data, with each image showing a hand-written digit at low resolution (28x28 px). The second dataset is FashionMNIST~\cite{xiao2017fashion}, which shares the same image size, number of classes, and structure of training and testing splits as the original MNIST. However, the FashionMNIST is a more challenging dataset that comprises ten classes of fashion products. We consider a shallow architecture for dynamic learning, which consists of four layers of a recurrent neural net: input 
layer, hidden layer, modulatory layer, and output layer. The plasticity in the hidden to output layer weights is regulated by using the modulatory layer. The modulatory layer corresponds to one of the eight meta-learning rules described by $\Psi$.

\begin{table}[]
    \centering
    \begin{tabular}{|c|c|}
    \hline
    Varaible & Search Space \\ \hline
      $\Psi$   &  \{GMR,MCR,NSCR,LMSR\\
                &  SLR,GUR,NSCoR,MOR\}\\ \hline
      $\alpha$   & [1e-03, 1e-00]\\ \hline
      $\beta_1$ & [1e-05, 1]\\ \hline
      $\beta_2$ & [1e-05, 1]\\ \hline
      $\beta_3$ & [1e-05, 1]\\ \hline
    \end{tabular}
    \caption{Search space considered to jointly optimize over the eight meta-learning rules and their parameters.}
    \label{tab:search_space}
    \vspace{-0.2in}
\end{table}
\begin{table}[]
    \centering
    \begin{tabular}{|c|c|c|} 
    \hline
    Dataset & Meta-Learning Rule  & Accuracy\\
    \hline
    MNIST   & LMSR & 0.903 \\
    \hline
    FashionMNIST & GMR  & 0.900 \\
    \hline
    \end{tabular}
    \caption{Best-performing meta-learning rule and corresponding testing accuracy with each dataset.}
    \label{tab:accuracy}
    \vspace{-0.3in}
\end{table}

The optimization search space consists of a categorical variable $\Psi$ that represents the choice of the meta-learning rule. In this work, we considered eight different rules, hence eight options for $\Psi$. The continuous parameters consist of $\alpha$, which represents the learning rate, and three continuous variables commonly defined for all the meta-learning rules---$\beta_1$, $\beta_2$, and $\beta_3$, as described in Section~\ref{sec:meta_def}. We note that not all the meta-learning rules have all the parameters, in which case only the subset of these parameters are used to evaluate the learning rule. However, the search algorithm is agnostic of this. The search space for all these parameters is shown in Table~\ref{tab:search_space}.

The optimization experiments are run by using an implementation of AMBS 
available in DeepHyper. The compute resources on Theta, a 11.69-petaflops leadership computing facility at Argonne, are used, where the experiments for each dataset are run on 128 Intel Xeon Phi (code-named Knights Landing) processor nodes.

\begin{figure}[h!]
    \centering
    (a) \\
    \includegraphics[width=1\linewidth]{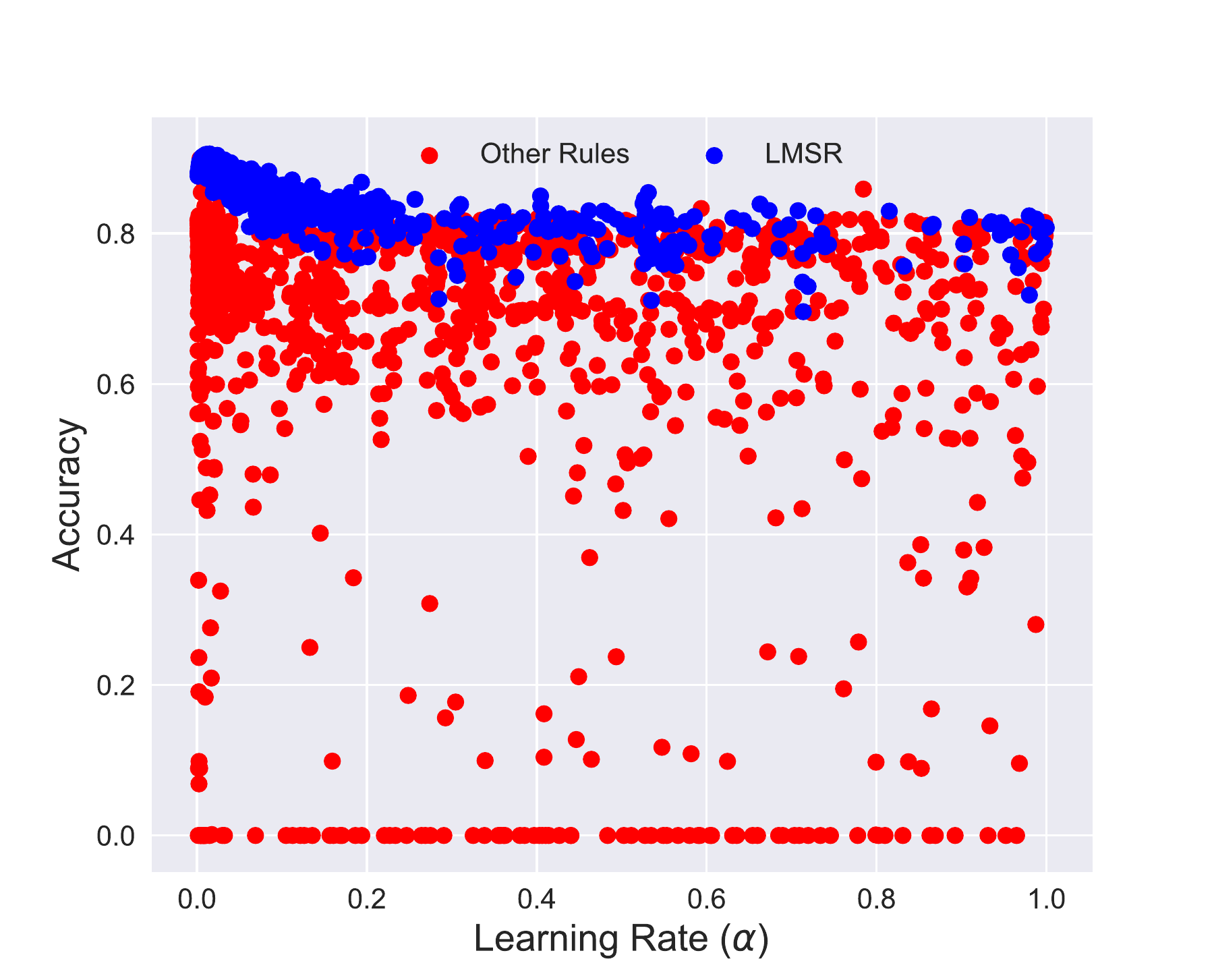}\\
    (b) \\
    \includegraphics[width=1\linewidth]{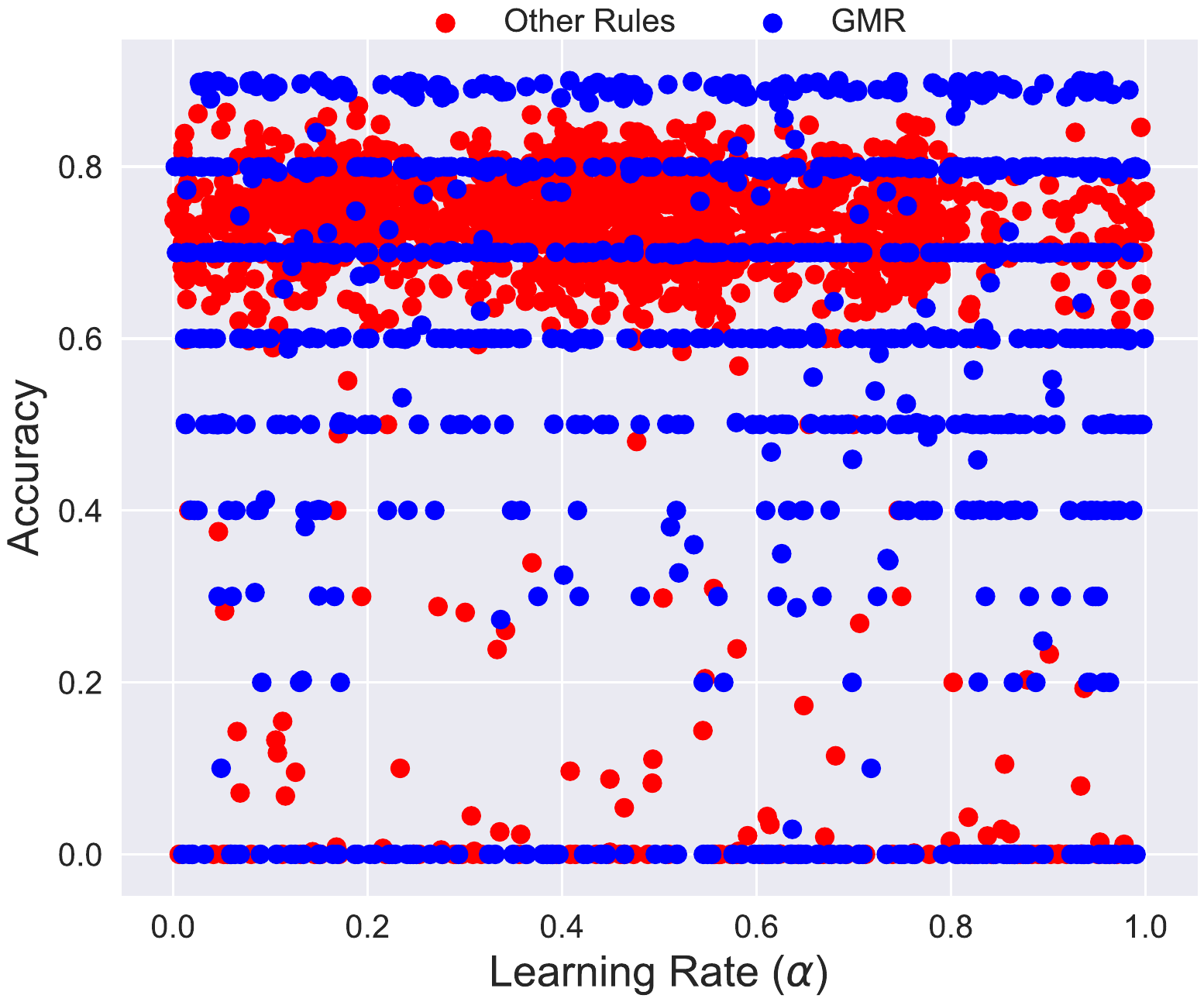}
    \caption{Testing accuracy as a function of the learning rate for all the evaluated configurations of the meta-learning rules where the evaluations corresponding to the best-performing rule is highlighted as opposed to all the other rules for (a) MNIST data and (b) Fashion-MNIST.}
    \label{fig:mnist_lr}
    \vspace{-0.4in}
\end{figure}

The shallow architecture has an input size of 784 (corresponding to images of size 28X28) and 10 outputs (corresponding to 10 classes) for both datasets. A total of 20,000 randomly selected images (from training data of 60,000) are used to train the model for each parameter choice, and the predictive accuracy is obtained on the testing data with 10,000 images for both datasets.  The best-performing meta-learning rule and its corresponding classification accuracy for both the MNIST and FashionMNIST datasets are shown in Table~\ref{tab:accuracy}. The corresponding scatter plots are shown in Figure~\ref{fig:mnist_lr}, where all the evaluations corresponding to the best-performing meta-learning rule are highlighted in blue while those corresponding to rest of the rules are highlighted in red. We note that the LMSR meta-learning rule with one active parameters ($\alpha$) works best for the MNIST dataset and obtains a parameter configuration that gives a classification accuracy of 0.903. On the other hand, the GMR rule has three active parameters $\alpha,\beta_1,\beta_2,\beta_3$ and provides the parameter configuration that achieves a classification accuracy of 0.9 for the FashionMNIST dataset, which is known to be more  complicated to learn than is MNIST.

Results show that LMSR meta-learning rule provides higher accuracy with the MNIST dataset for smaller learning rates, as is expected since the number of epochs used for training is low (0.33).  This dependence is not clear for the GMR rule performing best for the FashionMNIST data since its active search space is four dimensional compared with one dimensional for LMSR. We also observe that the $\beta_2$ is close to zero and $\beta_3$ is greater than $0.6$ for all the parameter configurations, producing an accuracy close to $0.9$ for GMR on FashionMNIST data.

\section{Conclusions}
We have modeled task-specific dynamic learning using insect brain-inspired mushroom body architecture implemented as a recurrent neural network, where the plastic synaptic weights are treated as layers in the network. The plasticity in output layer weights is controlled by the modulatory layer, which we model using meta-learning rules. This approach allows us to treat task-specific architecture optimization as the selection of the optimal meta-learning rule and its parameters. We employ a scalable, asynchronous model-based search approach to perform the optimization at scale on Theta, a leadership-class system at Argonne.  

We used two different machine learning benchmark datasets---MNIST and FashionMNIST---along with the proposed architecture and optimization to assess the predictive capability of the learning model. Because of the inherent differences in the two datasets, our proposed approach identifies different best-performing meta-learning rules for the datasets, thus emphasizing the need for task-specific dynamic learning.

In this work we have focused on a specific example to demonstrate our approach. Our goal is to apply this methodology to explore more complex, deeper architectures, including the presence of heterogeneous learning rules at different points of our network.

We also want to apply the same approach to the optimization of neuromorphic hardware, either to explore meta-learning in existing architectures such as Loihi or to help design novel robust and versatile architectures for edge-processing applications.

\section*{Acknowledgments}

This work was supported through the Lifelong Learning Machines (L2M) program from DARPA/MTO.
The material is also based in part by work supported by the U.S. Department of Energy, Office of Science, under contract DE-AC02-06CH11357.


%


\end{document}